\documentclass[letterpaper, 10 pt, conference]{ieeeconf}  

\usepackage{graphicx}
\usepackage{amsmath}
\usepackage{algorithm}%
\usepackage{algpseudocode}%
\usepackage{amssymb}

\IEEEoverridecommandlockouts                              

\title{\LARGE \bf
AnoF-Diff: One-Step Diffusion-Based Anomaly Detection for Forceful Tool Use
}

\author{Yating Lin$^{1}$, Zixuan Huang$^{1}$, Fan Yang$^{1}$, Dmitry Berenson$^{1}$
\thanks{$^{1}$University of Michigan
        {\tt\small yatinlin@umich.edu}}%
}

\begin{document}

\maketitle
\thispagestyle{empty}
\pagestyle{empty}

\begin{abstract}
Multivariate time-series anomaly detection, which is critical for identifying unexpected events, has been explored in the field of machine learning for several decades. However, directly applying these methods to data from forceful tool use tasks is challenging because streaming sensor data in the real world tends to be inherently noisy, exhibits non-stationary behavior, and varies across different tasks and tools.
To address these challenges, we propose a method, AnoF-Diff,  based on the diffusion model to extract force-torque features from time-series data and use force-torque features to detect anomalies. We compare our method with other state-of-the-art methods in terms
of  F1-score and Area Under the Receiver Operating Characteristic curve (AUROC) on four forceful tool-use tasks, demonstrating that our method has better performance and is more robust to a noisy dataset. We also propose the method of parallel anomaly score evaluation based on one-step diffusion and demonstrate how our method can be used for online anomaly detection in several forceful tool use experiments.

\end{abstract}

\section{INTRODUCTION}
\label{sec:intro}

As the development of robot sensing and machine learning technologies accelerates, multivariate time series analysis is becoming more and more critical in the robotics field. Robotic systems usually rely on current time-step sensor data for decision-making and control, which makes it possible to miss the potential temporal patterns over multiple time steps. Additionally, 
some sensor data, such as force-torque signals, require multiple time steps to capture dynamic behaviors and provide meaningful information. 

Due to the importance of temporal pattern recognition, several studies have explored using multivariate time series to analyze signal patterns across different domains, including robotics\cite{perez2015fast}, healthcare\cite{razaque2022anomaly}, and finance\cite{batres2015deep}. To achieve this, 
many time series algorithms have been developed, 
from traditional machine learning methods, such as Isolation Forest\cite{4781136}, to more recent deep learning approaches like Long Short-Term Memory (LSTM)\cite{nguyen2021forecasting}, Convolutional Neural Networks (CNNs)\cite{wu2023timesnettemporal2dvariationmodeling},  Transformers\cite{liu2024itransformerinvertedtransformerseffective}, and Diffusion\cite{yang2024surveydiffusionmodelstime}. These methods aim to capture both temporal and cross-dimensional dependencies from multivariate time series, enabling more accurate classification, forecasting, and anomaly detection.

Among these applications, multivariate time series anomaly detection (TSAD) \cite{Zamanzadeh_Darban_2024} has emerged as an important task in real-world scenarios, playing a vital role in ensuring the safety and reliability of robotic systems. However, obtaining anomaly labels is often challenging due to the rarity of anomalies and the effort of manual labeling. As a result, semi-supervised learning\cite{jiang2021semi} and unsupervised learning\cite{audibert2020usad} for anomaly detection are commonly employed, such as density-based methods\cite{zhang2018adaptive}, projection-based approaches\cite{blazquez2021review} and reconstruction-based method\cite{malhotra2016lstmbasedencoderdecodermultisensoranomaly}.
\begin{figure}
    \centering
    \includegraphics[width=\linewidth]{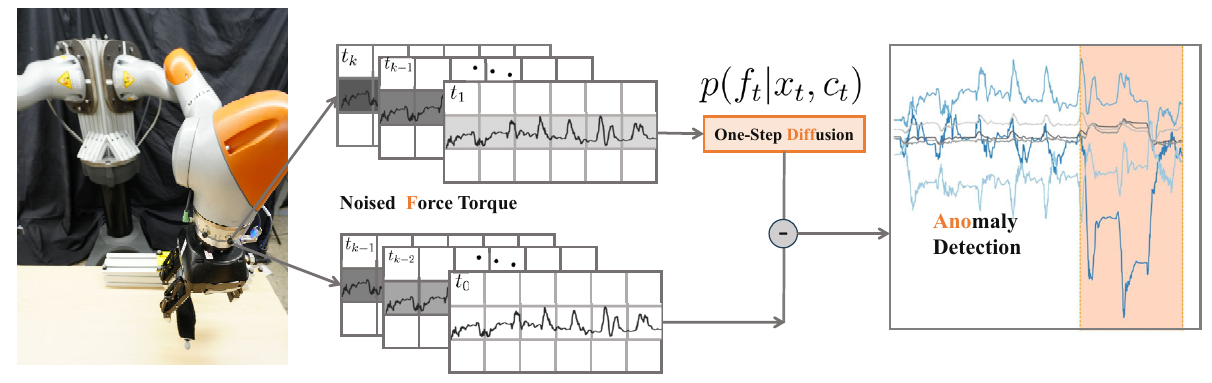}
    \caption{Parallel anomaly score evaluation by AnoF-Diff for forceful robotic manipulation. Adding noise to 6D force–torque signals collected from nut tightening tasks, denoising one-step using a conditional diffusion model to identify anomaly windows.}
    \label{fig:overview}
    \vspace{-5mm}
\end{figure}
Despite these advancements,  challenges remain when directly applying these methods to noisy robotics time series data.  The high dimensionality and noise in sensor data make it challenging to distinguish true anomalies from normal time series. Additionally, it is difficult to simultaneously capture the complex temporal dynamics and cross-dimensional dependencies inherent in the sensor data. Addressing these challenges is essential to ensure the reliability and safety of robotic systems operating in dynamic and uncertain environments for forceful tool-use tasks.

In this paper, we consider the problem of determining anomalies when a robot has executed a forceful tool-use task, such as tightening a nut or prying apart two pieces of wood. We collect multivariate time series data consisting of the robot gripper pose and the end-effector force-torque from these tool-use tasks.
Our goal is to create a method that uses this data to inform a control policy about whether to keep going or stop as a result of task presence of an anomaly. To accomplish this, we propose a diffusion model that denoises the state-dependent features, specifically force-torque features, to reconstruct the multivariate time series inputs. The reconstruction loss is then used as an anomaly detection score to identify abnormal inputs. Considering real-world anomaly detection scenarios, we also introduce a parallel anomaly score evaluation method based on one-step diffusion to efficiently assess multiple denoising outcomes simultaneously. This parallel evaluation strategy improves detection speed without sacrificing performance, which is critical for enabling real-time anomaly monitoring in dynamic and noisy environments.
We test our methods on several forceful tool use manipulation tasks and compare with state-of-the-art baseline methods.
 
The main contributions of our paper are:
\begin{itemize}
    \item A framework that uses a diffusion model to denoise state-dependent features for multivariate time series anomaly detection.
    \item A parallel anomaly score evaluation method based on one-step diffusion denoising, enabling efficient real-time performance.
    \item Evaluation on real-world forceful tool use experiments, both offline and online, which demonstrates that our method outperforms leading baseline methods.
\end{itemize}


\section{Related Work}
\label{sec:citations}

\textbf{Force-Aware Robotic Manipulation.} Force-aware robotic manipulation refers to the ability to perceive and control force during interaction with objects and the environment. This capability is important to achieve dexterity\cite{he2024foar}, precision\cite{luo2021learning}, and safety\cite{wei2024ensuringforcesafetyvisionguided} in robotic manipulation.  In particular, planning with both visual and force feedback is essential for complex tool-use tasks \cite{holladay2019force}. Reinforcement learning\cite{luo2021learning}\cite{inoue2017deep} \cite{noseworthy2025forgeforceguidedexplorationrobust} and learning from demonstrations\cite{lee2015learning}
have been widely explored to allow robots to acquire force-aware manipulation skills.  However, these approaches often struggle with generalization to unseen environments. While diffusion models\cite{ho2020denoisingdiffusionprobabilisticmodels} with strong generative capabilities and robustness in modeling complex multi-modal distributions have gained more attention in the robotic manipulation area. Several studies have explored the integration of force sensing with diffusion models to enhance the manipulation performance of forceful tool-use tasks\cite{he2024foarforceawarereactivepolicy}\cite{kang2025roboticcompliantobjectprying}\cite{wu2025tacdiffusionforcedomaindiffusionpolicy}.
But among all these works, they mainly focus on optimizing task performance but do not explicitly consider the challenge of detecting out-of-distribution/anomaly events during execution in an uncertain and dynamic environment.

\textbf{Time Series Anomaly Detection}
Considering the importance of monitoring anomalies in robotic systems, time series anomaly detection has been widely adopted, as robotic sensor data are inherently structured as multivariate time series. The existing TSAD methods typically fall into three major categories\cite{boniol2024divetimeseriesanomalydetection}, distance-based methods, density-based methods, and learning-based methods. The distance-based method\cite{8970686} identifies anomalies by measuring how far a data point deviates from its neighbors.   Density-based methods\cite{4781136} treat the time series as a more complex architecture instead of single values,  while both of them struggle with the curse of dimensionality.  To overcome these challenges, learning-based TSAD approaches\cite{Zamanzadeh_Darban_2024} have been developed, which can generally be categorized into forecasting-based and reconstruction-based methods. Forecasting-based methods predict the next time step or window based on the observed time series data. Anomalies are then identified by comparing the difference between the predicted window and the actual observed values. Recent work in this vein has also integrated graph structure with transformers \cite{Chen_2022_IOT}, achieving state-of-the-art performance. However, forecasting-based methods often face challenges in handling rapidly and continuously changing time series. In contrast, the reconstruction-based method is based on reconstruction errors to detect anomalies, which are particularly suited for capturing collective and contextual anomalies. Examples of reconstruction-based methods include Convolutional and LSTM-based autoencoders (AE) \cite{hasan2016learningtemporalregularityvideo}\cite{MSCRED}, LSTM-based variational autoencoders (VAEs)\cite{10.1145/3485447.3511984}\cite{zhang2019semisupervisedlearningbearinganomaly}\cite{park2017multimodalanomalydetectorrobotassisted}, and transformer-based reconstruction models, Anomaly Transformer\cite{xu2022anomalytransformertimeseries}. Recently, diffusion models have gained attention for their success in robotic manipulation tasks and have also demonstrated strong potential in time series anomaly detection\cite{pintilie2023timeseriesanomalydetection}\cite{chen2023imdiffusionimputeddiffusionmodels}\cite{livernoche2025diffusionmodelinganomalydetection} by modeling noisy data distributions and capturing the underlying structure of multivariate time series.

However, previous work mainly focuses on capturing temporal relationships within the time series, without considering the cross-dimensional dependencies between different input modalities, such as the state dependency of sensor measurements. In our work, we build on prior research in diffusion models for robotic manipulation and TSAD. Our work focuses on leveraging temporal patterns from multivariate time series while incorporating force–torque information to predict anomalies for forceful tool-use manipulation tasks. In addition, instead of evaluating only on offline anomaly datasets, we consider deploying our anomaly detection model in online scenarios. To support this, we propose a parallel anomaly score evaluation strategy that enables efficient and real-time anomaly detection during robot execution.

\begin{figure*}[ht]
\vspace{3mm}
    \centering \includegraphics[width=0.9\linewidth]{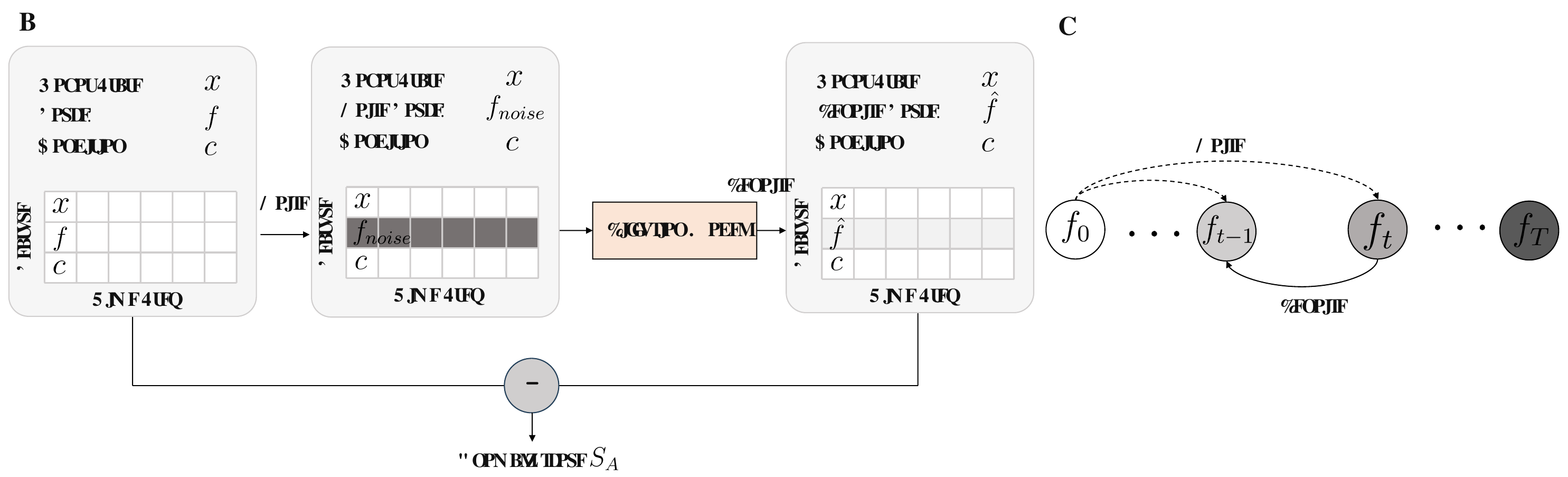}
    \caption{Method Overview: (a) Structure of AnoF-diff. The input features are divided into three categories: the robot state (e.g., position, orientation), the sensor measurements (e.g., force, torque), and conditional variables relevant to the task. (b) Illustration of one-step denoising for anomaly detection, where the model directly denoise from $f_t$ and compares the results with $f_{t-1}$ steps}
    \label{fig:one_step_denoise}
    \vspace{-3mm}
\end{figure*}

  
\section{Problem Statement}
\label{sec:problem_statement}
\textbf{Definition}
In this paper, we consider multivariate time series signals $X_T = [x_1,x_2,\cdots,x_T]$ collected from complete task execution. $x_i$ is the state-sensor data collected at time step $i$. $x_i \in R^N$, where $N$ is the dimension of state-sensor data, such as the gripper pose, end-effector force-torque, and object pose. $X_T\in R^{T\times N}$, where $T$ is the total number of time steps in the multivariate time series. During training, we use the sliding window to chop the multivariate time series data $X$. The length of the sliding window is defined as $L_w$, then $X^i_w = [x_i,x_{i+1},\cdots,x_{i+L_w-1}] $ represents the window of data starting at time step $i$ with window length $L_w$. 
We define all  windows sampled from $X_T$ as the training dataset $\{D_{train}|X^i_w, i\in[0,T-L_w+1]\}$. Any input $X_w$ in the training data set $D_{train}$ is defined as in-distribution. Given in-distribution $X_w$,
for different tasks, each $X_w$ has the corresponding label $L$, specifying whether the task succeeded in the given window. Otherwise, we consider $X_w \notin D_{train}$ to be an abnormal time series window. For out-of-distribution inputs $X_w$, we define $\delta: X\rightarrow \{0,1\}$ to evaluate whether this data is anomolous, where $\delta(X_w)=1$ indicates an anomaly and $\delta(X_w)=0$ indicates in-distribution inputs. Here $S_A(\cdot)$ is defined as the anomaly score function, $\tau$ is the anomaly threshold.
\begin{equation}
\delta = 
\begin{cases}
1 & \text{if } S_A(\mathbf{X}_w) > \tau \\
0 & \text{otherwise}
\end{cases}
\end{equation}

\textbf{Evaluation Protocol}
Precision, recall, and F1-score are fundamental evaluation metrics for anomaly detection tasks, especially with imbalanced datasets. Precision measures the proportion of true positive predictions out of all positive predictions predicted by the model and is defined as:
\begin{equation}
\begin{aligned}
\text{Precision} &= \frac{\text{TP}}{\text{TP} + \text{FP}}, \quad
\text{Recall} &= \frac{\text{TP}}{\text{TP} + \text{FN}}
\end{aligned}
\end{equation}
where \( \text{TP} \) represents true positives and \( \text{FP} \) represents false positives. Recall, on the other hand, quantifies the ability of the model to capture positive samples.
where \( \text{FN} \) represents false negatives. F1-score is a harmonic mean of precision and recall, offering a metric that balances between measures of precision and recall, which is defined as:
\begin{equation}
\text{F1} = 2 \cdot \frac{\text{Precision} \cdot \text{Recall}}{\text{Precision} + \text{Recall}}.
\end{equation}

Based on the two approaches for selecting the anomaly threshold, we define two evaluation metrics: calibrated $F1$ score, $F1_c$, and the best $F1$ score, $F1_{best}$. $F1_c$  measures the $F1$ score when the threshold is determined using a calibration dataset, where the calibration dataset consists of normal trajectories from real-world execution. $F1_{best}$, on the other hand, measures the $F1$ obtained by selecting the optimal threshold that maximizes the $F1$ score on the test set, where the test dataset includes both normal and abnormal trajectories and corresponding labels.
In addition to these two metrics, we also use the Area Under the Receiver Operating Characteristic curve (AUROC) to evaluate the model’s discriminative ability across all possible thresholds. $F1_c$ and $F1_{best}$ capture threshold-specific performance, and AUROC provides a threshold-independent measure that reflects the overall quality of the predictions.


\section{Method}
\subsection{Denoising Diffusion Probabilistic Models}
Denoising Diffusion Probabilistic Models (DDPMs)
{\cite{ho2020denoisingdiffusionprobabilisticmodels}}  represents a state-of-the-art class of generative models that is capable of approximating complex and multimodal data distributions. DDPMs consist of two processes. The forward process adds Gaussian noise to input data as shown in  eq.\ref{eq:q_xt_x0}.
\begin{equation}
q(x_t|x_0) = \mathcal{N}\left(x_t |  \sqrt{\bar{\alpha}_t}x_0, (1 - \bar{\alpha}_t)\mathbf{I} \right)
\label{eq:q_xt_x0}
\end{equation}
\begin{equation}
x_t = \sqrt{\bar{\alpha}_t}x_0  + \sqrt{1 - \bar{\alpha}_t}  \epsilon_t, \quad \epsilon_t \sim \mathcal{N}(0, \mathbf{I})
\label{eq:x_t}
\end{equation}
Here, $x_0$ is the original input and $t$ is noise time steps. The term $\bar{\alpha}_t$ represents the cumulative product of noise scaling coefficients up to timestep $t$, which can be written as $\bar{\alpha}_t = \prod_{s=1}^{t} \alpha_s = \prod_{s=1}^{t} (1 - \beta_s)$. And $\beta$ is the noise schedule that controls the amount of Gaussian noise added at each diffusion step. The reverse process starts from a standard normal distribution and iteratively denoises the sample to recover the original data distribution. The sampling equation shown in eq.\ref{eq:sampling} describes how to generate the denoised sample at each timestep during the reverse diffusion process.
\begin{equation}
p_\theta(x_{t-1} \mid x_t) := \mathcal{N}\left(x_{t-1} | \mu_\theta(x_t, t), \Sigma_\theta(x_t, t)\right)
\end{equation}
\begin{equation}
\mu_\theta(x_t, t) = \frac{1}{\sqrt{\alpha_t}} \left( x_t - \frac{1 - \alpha_t}{\sqrt{1 - \bar{\alpha}_t}} \epsilon_\theta(x_t, t) \right)
\label{eq:sampling}
\end{equation}
Here $\epsilon_\theta$ is a trained diffusion model to predict the noise component at time steps $t$. By using eq.\ref{eq:sampling}, we can recover the original input from the noise sample.
Compared to conventional autoencoder-based approaches, such as VAEs and AEs, the diffusion model can learn a more complex data distribution by corrupting the input with noise and training a denoising model to denoise. 

\subsection{Anomaly Force 
Diffusion (AnoF-diff)}
Based on the forward and reverse processes of DDPMs, AnoF-Diff introduces a conditional masking strategy that selectively diffuses only the force-torque components of the input while preserving the robot state and task-specific conditions as auxiliary inputs as Fig. \ref{fig:one_step_denoise} (a) shows.

Instead of applying diffusion to the entire multivariate time-series input, this approach learns the conditional probability distribution 
$p(f{|} x,c)$, where 
$f$ denotes the observed force-torque signal  and
$x$ represents the state of the robot, and $c$ represents the conditional information. In principle, this approach can also be generalized to any other state-dependent time series measurements. This formulation allows the model to learn the underlying structure of force-torque dynamics conditioned on state-related features. In contrast to prior anomaly detection methods that treat all input dimensions uniformly and primarily focus on modeling temporal relations, our proposed method explicitly distinguishes between robot state variables and sensory feedback signals.  This design enables more informed and interpretable anomaly detection that accounts for both contextual state and sensor-specific behavior.

To capture temporal dependencies within the time-series data, we employ a 1D U-Net as the architecture of AnoF-Diff. The 1D U-Net is a symmetric encoder-decoder architecture with skip connections. The encoder captures temporal features of noise inputs through successive 1D convolutions and downsampling. The decoder predicts the corresponding noise based on the noisy input and the corresponding diffusion timestep. Skip connections directly link encoder layers to corresponding decoder layers, preserving detailed temporal features that could be lost during downsampling.

\subsection{Parallel Anomaly Score Evaluation}
At inference time, we propose a parallel anomaly score evaluation strategy to accelerate computation, enabling real-time applicability of the proposed method. To compare the original input force time series with its denoised counterpart, the traditional DDPMs approach requires an iterative denoising process over $n$ steps, which may hinder real-time application for anomaly detection. Therefore, instead of going through the entire denoising process to reconstruct the original time series input, we directly evaluate the anomaly score based on the one-step denoise process, as Fig.\ref{fig:one_step_denoise}(b) shows.  $f$ represents the force-torque time series input. Based on eq.\ref{eq:x_t}, given the first steps input, we can add noise to any $t$ noise time steps. With the trained diffusion model, by applying eq.\ref{eq:sampling}, we are able to get the predicted $\hat{f}_{t-1}$. Therefore, the anomaly score $S_A$ can be written as the following equation, here $D_1(\cdot)$ means one-step denoising process.

\begin{equation}
\begin{aligned}
    S_A &= \left\| f_{t-1} - D_1(f_t) \right\|_2 , \quad
      \bar{S}_A = \frac{1}{K}\sum_i^K||f_{t_i -1} - D_1(f_{t_i})||_2 
\end{aligned}
\label{eq:parallel}
\end{equation}
For in-distribution inputs, the one-step denoising prediction closely approximates the ground truth, whereas for anomalous inputs, the discrepancy between the prediction and the true signal tends to be larger.
Based on the one-step denoising comparison, we can parallelize the process across $K$ samples to get the average anomaly score as  eq.\ref{eq:parallel} shows. Here, K samples correspond to adding different noisy timesteps $t$ by the forward diffusion process independently. The use of K samples is a design choice intended to capture gradient features across varying noise levels in the reverse diffusion. The detail parallel one-step denoising as the Alg. \ref{alg: Ano-Diff} shows.
Unlike the iterative denoising procedure in traditional DDPMs, this approach enables simultaneous parallel evaluation, significantly improving inference efficiency.
\begin{algorithm}
\caption{Parallel One-Step AnoF-Diff with K Samples}
\label{alg: Ano-Diff} 
\begin{algorithmic}
\State \textbf{Given:} Multivariate time series signal $X \in \mathbb{R}^{T \times N}$. Fixed window length $L$. Anomaly threshold $\tau$. Diffusion model $D_\theta(\cdot)$.
Force/torque indices $\mathcal{F}$.
\For{Anomaly Detection}
    \State Extract the last $L$ time steps: $X^{w} \gets X[T - L : T]$
    \State Tile $X^{w}$ to shape $K \times L \times N$: $X^w \gets \text{Repeat}(X^w, K)$
    \State Sample batch of diffusion steps $\mathbf{t} = [t_1, \dots, t_K]$ and noise $\mathbf{\epsilon} = [\epsilon_1, \dots, \epsilon_K]$ independently
    \State Forward Diffusion: add batch noise $\epsilon$ to force/torque dimensions $\mathcal{F}$ get noised window ${X_t^w} = [X_{t_1}^w, \dots, X_{t_K}^w]$ and ${X_{t-1}^w} = [X_{t_1-1}^w, \dots, X_{t_K-1}^w]$
    \State One-Step Reverse Diffusion: \\
    \hspace*{1.5em}
    $\hat{X}_{t-1}^w = D_\theta(X_t^w,\mathbf{t})  = [\hat{X}_{t_1-1}^w, \dots, \hat{X}_{t_K-1}^w]$ 
    \State Anomaly Score:  \\
    \hspace*{1.5em}
    $S_A^w =||{X}_{t-1}^w-\hat{X}_{t-1}^w||_2  = [S_{t_1-1}^w, \dots, S_{t_K-1}^w]$ 
    \State Average Anomaly Score: $\bar{S}_A^w = \frac{1}{K}\sum_i^KS_{t_i -1}^w $

    \If{$\bar{S}_A^w > \tau$}
        \State Break and Stop Robot
    \Else
        \State Keep detection and Update $X$
    \EndIf
\EndFor
\end{algorithmic}
\end{algorithm}

\subsection{Training Process of AnoF-Diff}
In this section, we discuss how we pre-process and label the multivariate time series data collected from the robot sensor to improve the model performance. As defined in Sec. \ref{sec:problem_statement}, after we collect the robot gripper pose, object pose and end effector force-torque time-series data, we use a sliding window to sample the whole time-series data.  The length of a sliding window $L_w$ is a hyperparameter selected by the user. Sampling fixed length windows facilitates parallel training. 
Before inputting the time series window  $X_w$ into the model, we normalize the dataset by Min-Max scaling. Additionally, for any quaternion rotation inputs in the time series data, due to discontinuity, we convert them to a 6D rotation representation, which is more suitable for learning \cite{zhou2020continuityrotationrepresentationsneural}. The training loss function can be expressed as follows. The input to the model is a noisy force time series $x_t$ along with the corresponding noise timestep $t$, and the output is the predicted noise $\epsilon_\theta$. The objective is to minimize the difference between the predicted noise and the true noise $\epsilon$ that was added according to the noise schedule. 
\begin{equation}
\mathcal{L}_{\text{diff}} = \mathbb{E}_{x_0, \epsilon, t} \left[ \left\| \epsilon - \epsilon_\theta(x_t, t) \right\|_2^2 \right]
\end{equation}


\section{Experiments}
\label{sec:result}
\begin{figure*}
\vspace{3mm}

    \centering
    \includegraphics[width=0.9\linewidth,
    height=0.4\linewidth]{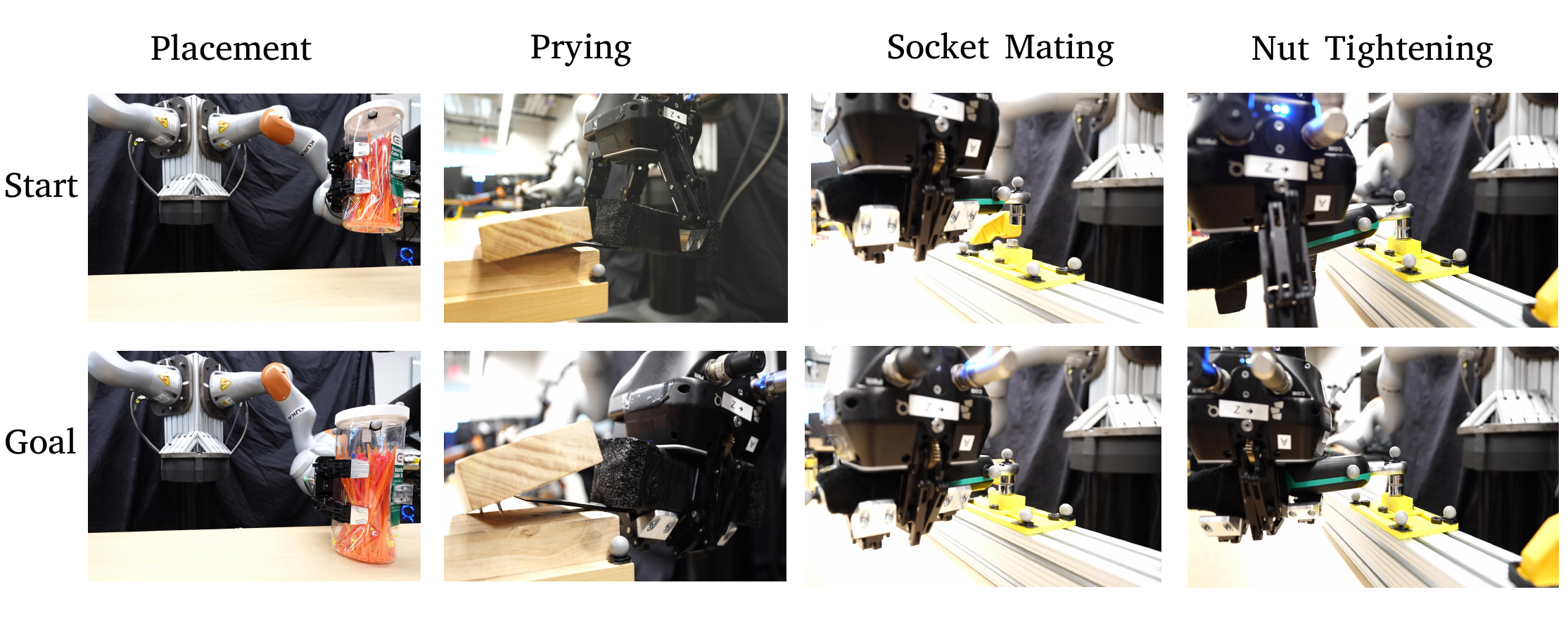}
    \caption{Normal data collection (1) Placement: the robot stably places a plastic bottle on the table. (2)Prying: The robot uses a tool to separate two wooden planks that are fixed together with two nails (3) Socket mating: The robot aligns the socket wrench with a nut, which is fixed to the table. (4) Nut Tightening: The robot tightens the nuts using the socket wrench, assuming the socket is already mated with the nut.}
    \label{fig:real_experiment}
\vspace{-3mm}

\end{figure*}
In the experiments section, we first introduce the baseline and ablation methods we compared with. Then we will explain the tasks that are used to verify our method and how we collect normal training data from each task. In the results part, we first present the anomaly prediction accuracy of our method based on the traditional DDPM denoising process. We then compare the performance of the iterative denoising procedure with our proposed parallel denoising approach, demonstrating that parallel denoising achieves comparable performance while improving efficiency. Finally, we describe how our trained model is deployed in the online setting for real-time anomaly detection.
\subsection{Baselines}
We evaluate our method against several baseline models, including the classical method, iForest\cite{4781136}, AutoEncoder based methods, Conv-AE\cite{hasan2016learningtemporalregularityvideo}, LSTM-AE\cite{malhotra2016lstmbasedencoderdecodermultisensoranomaly}, VAE\cite{10.1145/3485447.3511984}, Conv-VAE\cite{zhang2019semisupervisedlearningbearinganomaly}, LSTM-VAE\cite{park2017multimodalanomalydetectorrobotassisted}, and state-of-art methods, Anomaly Transformer~\cite{xu2022anomalytransformertimeseries},  and ImDiffusion~\cite{chen2023imdiffusionimputeddiffusionmodels}.
Anomaly Transformer\cite{xu2022anomalytransformertimeseries} is one of the state-of-the-art anomaly detection baselines, utilizing a transformer architecture combined with prior association and series association to reconstruct multivariate time series signals and perform anomaly detection. 
ImDiffusion\cite{chen2023imdiffusionimputeddiffusionmodels} represents recently proposed anomaly detection baselines based on diffusion model, 
employing grating masking and imputation strategies to improve the reconstruction of multivariate time series and facilitate anomaly detection. 
We also consider one ablation of our method as a baseline, which directly denoises on the whole dimension instead of the force-torque dimension.
 \subsection{Data collection}
All experiments were performed in the real world. We used a 7-DoF KUKA LBR iiwa robotic arm to perform the tasks, collecting time-series data of the gripper pose and end-effector force-torque. A motion capture system is used to determine the object's pose and track its movement during task execution. We evaluated our method across four representative manipulation tasks: placement, prying, socket mating, and nut tightening. For each task, we construct the training dataset by randomly sampling from different trials, resulting in a total of 20,000 multivariate time-series windows. The start and goal conditions for normal trajectories are as Fig.\ref {fig:real_experiment} shows. The calibration dataset is also obtained by random sampling from the normal trajectories, excluding them from training. It consists of 7,000 multivariate time-series sliding windows and is used to calculate the anomaly threshold for offline evaluation. The abnormal test dataset for each task is generated by introducing perturbations in arbitrary directions at the robot’s end-effector during task execution, simulating scenarios where the robot becomes stuck due to external obstructions. In total 5,000 multivariate time-series windows are sampled, with random sampling yielding a normal-to-anomaly ratio of 2:3.

For the placement task, the robot starts from a random pose above the table and aims to place an object at the table steadily. Inputs include the gripper pose relative to the object and force-torque time series. In the prying task, the robot grasps a tool near the connection point of two nailed wooden blocks and applies a pre-defined separating force. We use the gripper pose relative to the under-block and the force-torque data as inputs. In the socket mating task, the robot begins with a pose-offset socket wrench near the nut and must achieve proper alignment for engagement. We collect trajectories, including gripper-to-bolt pose, force-torque data, and the bolt pose in the robot frame as a conditional input. For nut tightening, the robot starts with a correctly mated socket wrench and applies torque to fasten the nut within a specified range. The data inputs are identical to those used in the socket mating task.

\subsection{Experiments Results}
\begin{table*}[ht]
\vspace{4mm}
    \centering
    \scalebox{0.85}{
    \begin{tabular}{c|ccc|ccc|ccc|ccc}
        \hline
        \multicolumn{1}{c|}{\textbf{Method}} 
        & \multicolumn{3}{c|}{\textbf{Placement}} 
        & \multicolumn{3}{c|}{\textbf{Prying}}
        & \multicolumn{3}{c|}{\textbf{Socket Mating}}
        & \multicolumn{3}{c}{\textbf{Nut Tightening}}
        \\
        \cline{2-13}
        & $F1_c$ & $F1_{best}$ & AUROC
        & $F1_c$ & $F1_{best}$ & AUROC
        & $F1_c$ & $F1_{best}$ & AUROC
        & $F1_c$ & $F1_{best}$ & AUROC
        \\
        \hline
        iForest\cite{4781136}
        & 0.567 & 0.660 & 0.674
        & 0.883 & 0.928 & 0.906
        & 0.497 & \underline{0.808} & 0.666
        & 0.063 & 0.901 & 0.516\\ 
        Conv AE\cite{hasan2016learningtemporalregularityvideo}   
        & 0.874 & 0.913 & 0.969
        & 0.801 & 0.963 & \underline{0.989} 
        & 0.469 & 0.709 & 0.769 
        & 0.925 & 0.933 & \textbf{0.983} \\
        
        LSTM AE\cite{malhotra2016lstmbasedencoderdecodermultisensoranomaly}           
        & 0.800 & 0.827 & 0.926
        & \underline{0.911} & 0.938 & 0.981
        & 0.069 & \textbf{0.840} & \underline{0.888}
        & 0.328 & 0.847 & 0.908 \\
        
        VAE\cite{10.1145/3485447.3511984}   
        & 0.703 & 0.777 & 0.819
        & 0.879 & 0.896 & 0.897
        & 0.112 & 0.571 & 0.624
        & 0.678 & 0.728 & 0.831 \\

        Conv VAE \cite{zhang2019semisupervisedlearningbearinganomaly}         
        & \underline{0.910} & \underline{0.922} & \underline{0.974}
        & 0.818 & 0.962 & 0.987
        & 0.617 & 0.681 & 0.780
        & 0.675 & 0.690 & 0.770 \\

        LSTM VAE\cite{park2017multimodalanomalydetectorrobotassisted}          
        & 0.886 & 0.906 & 0.935
        & 0.883 & \textbf{0.969} & 0.985
        & 0.485 & 0.637 & 0.636
        & \underline{0.907} & 0.920 & \underline{0.982} \\

        Anomaly Transformer\cite{xu2022anomalytransformertimeseries} 
        & 0.822 & 0.915 & 0.852
        & 0.792 & 0.952 & 0.825
        & \underline{0.647} & 0.756 & 0.631
        & 0.905 & \underline{0.934} & 0.914 \\

        lmdiffusion\cite{chen2023imdiffusionimputeddiffusionmodels}        
        & 0.819 & 0.897 & 0.971
        & \textbf{0.924} & 0.955 & 0.988
        & 0.428 & 0.717 & 0.823
        & 0.282 & 0.682 & 0.724 \\
        
        Diffusion Ablation 
        & 0.903 & 0.910 & \underline{0.974}
        & 0.892 & 0.912 & 0.963
        & 0.250 & 0.582 & 0.626
        & 0.229 & 0.572 & 0.565 \\
        
        Ours (AnoF-Diff) 
        & \textbf{0.926} & \textbf{0.929} & \textbf{0.984}
        & 0.907 & \underline{0.968} & \textbf{0.992}
        & \textbf{0.786} &{0.797}   & \textbf{0.909}
        & \textbf{0.932} & \textbf{0.936} & 0.971\\
        \hline
    \end{tabular}}
    \caption{Anomaly Detection Comparison of methods for Placement, Prying, Socket Mating and Nut Tightening tasks. Bold numbers indicate the best performance, while underlined numbers represent the second-best performance for each metric.}
    \label{tab:comparison}
    \vspace{-5mm}
\end{table*}

\textbf{Anomaly Detection}
 For anomaly detection, we used a selected threshold on reconstruction error to determine if an input is anomalous. We use the $F1 $ score and AUROC as the primary metric. A higher $F1$ score and AUROC indicate superior performance in identifying abnormal inputs. We use two ways to preselect the threshold as the sec.\ref{sec:problem_statement} mentioned. One threshold is selected as one-half standard deviation above the mean reconstruction loss computed from the calibration dataset, which we call $F1_c$. Another $F1_{best}$ is calculated by searching for the optimal threshold based on labeled abnormal test dataset, which represents the best performance the model is able to achieve, but note that this is not realistic as it requires knowing the dataset of abnormal data \textit{a priori}. Each model’s anomaly threshold is determined based on its own prediction errors. We evaluate our method on four manipulation tasks that involve force-torque information and tool-use as Fig. \ref{fig:real_experiment} shows.
Here, we present the anomaly detection results in Table \ref{tab:comparison} for placement, prying, socket mating, and nut tightening task. 

From the result table, the first observation is that among all tasks, the socket mating task exhibits lower $F1$ scores compared to the others. This can be attributed to two primary factors. First, the data distribution for the socket mating task is more complex than that of the other tasks. In addition to successful mating trajectories, the dataset also includes numerous failed mating attempts due to
the task difficulty. Second, some of these failed trajectories exhibit large force-torque magnitudes, making it difficult to distinguish between normal failed attempts and externally induced anomalies, thereby complicating the anomaly detection process. From the results, it can be observed that our method achieves superior performance compared to all baselines with respect to $F1_c$ and AUROC. In terms of $F1_{best}$ score, it is also comparable to the best value. In comparison to the diffusion ablation method, our proposed model better captures the conditional distribution of force-torque signals given the robot state and conditions, improving its ability to distinguish anomalies more effectively along the force-torque dimensions. Although the LSTM-AE model gets the highest $F1_{best}$ score, it also exhibits the lowest $F1_c$, suggesting that under the same calibration conditions,  the LSTM-AE model captures the least information regarding the optimal anomaly threshold,  limiting its applicability in real-world online scenarios.
\begin{table}[ht]
    \centering
    \scalebox{0.85}{
    \begin{tabular}{c|ccc|ccc}
        \hline
         \multicolumn{1}{c|}{\textbf{Task}}  &
        \multicolumn{3}{c|}{\textbf{AnoF-Diff (Iterative)}} & \multicolumn{3}{c}{\textbf{AnoF-Diff(Parallel)}} \\
        \cline{2-7}
        & $F1_c$ & $F1_{best}$ & AUROC
        & $F1_c$ & $F1_{best}$ & AUROC\\
        \hline
        Placement 
        & 0.926 & 0.929 & 0.984
        & 0.925 & 0.926 & 0.967\\
        Prying 
        & 0.907 & 0.968 & 0.992
        & 0.908 & 0.939 & 0.965\\
        Socket Mating 
        & 0.786 & 0.797 & 0.909 
        & 0.862 & 0.881 & 0.959\\
        Nut Tightening 
        & 0.932 & 0.936 & 0.971 
        & 0.917 & 0.941 & 0.974\\
        \hline
    \end{tabular}
    }
    \caption{Anomaly Detection Comparison of iterative denoising and parallel denoising for AnoF-Diff.}
    \label{tab:comparison_parallel}
\vspace{-5mm}
\end{table}
The second observation is that baseline methods exhibit considerable variability in performance across different tasks, indicating limited generalization capabilities. In contrast, our proposed method maintains consistent performance across all four tasks due to consider the force-torque feature. Although it does not achieve the highest scores in the prying tasks, it still outperforms the majority of baseline models, demonstrating stronger robustness and adaptability to varying task conditions. 

In addition to the overall comparison, we further analyze the performance of state-of-the-art baseline methods to better understand their strengths and limitations across different tasks.
Anomaly Transformer~\cite{xu2022anomalytransformertimeseries} exhibits the second-best robustness among all evaluated baselines. Considering the transformer architecture and the training strategy that incorporates prior association and cross-series association, Anomaly Transformer is better than other methods to capture both temporal dependencies and Gaussian prior dependencies, contributing to its strong robustness across tasks. However, since the vanilla Transformer architecture does not inherently capture inter-dimensional dependencies, this limitation may contribute to the superior performance of our proposed method compared to the Anomaly Transformer.
In terms of ImDiffusion,
despite it achieving the highest score on the prying task, 
ImDiffusion~\cite{chen2023imdiffusionimputeddiffusionmodels} exhibits lower performance on the remaining tasks. A possible reason is that its grating masking mechanism primarily focuses on local imputations, which limits its ability to capture collective anomalies. Additionally, the grated distribution may be more complex than the conditional distribution of force-torque signals or even the original data distribution, potentially leading to degraded performance across our tasks.

\begin{table}[t]
\centering
\renewcommand{\arraystretch}{1.2}
\begin{tabular}{l|c}
\hline
\textbf{One batch in GPU 4090} & \textbf{Speed (s)} \\
\hline
Anomaly Transformer & 0.0101 \\
Imdiffusion & 0.1834 \\
AnoF-Diff (Iterative) & 0.1936 \\
AnoF-Diff (Parallel) & \textbf{0.0023} \\
\hline
\end{tabular}
\caption{\textsc{Inference speed comparison across different methods.}}
\label{tab:vertical_speed_comparison}
\vspace{-10mm}
\end{table}

\textbf{Parallel Denoising}
In this section, we present a comparison between iterative denoising and parallel denoising strategies for anomaly detection across four manipulation tasks. Here, we chose the 10 noise time steps as the parallel sample to denoise and the chosen steps are $[1,10,20,30,\cdots,90]$.
As shown in Table~\ref{tab:comparison_parallel}, we report the calibrated $F1_c$, the best $F1$ score $F1_{best}$, and AUROC for each task under both denoising approaches.

The results demonstrate that parallel denoising achieves comparable or even superior performance relative to iterative denoising in some tasks, while offering significant advantages in computational efficiency. The superior performance of parallel denoising observed in the socket mating task may also be attributed to the complexity of the underlying learning distribution. The iterative denoising process proceeds from the final time steps back to the initial ones. During multiple diffusion steps, prediction errors can accumulate. Additionally, when the data exhibit a multi-modal distribution, even normal samples may result in denoising outputs that deviate from the original input. However, for parallel one-step denoising, the model predicts the noise in a single step, thereby reducing the accumulation of errors across diffusion steps, which improves reconstruction accuracy for data with complex multi-modal distributions.
We compared the running speeds of Anomaly Transformer, Imdiffusion, and AnoF-Diff (Iterative and Parallel). Table \ref{tab:vertical_speed_comparison} shows that the parallel denoising of AnoF-Diff achieves the best efficiency. The magnitude of the speedup depends on the number of iterative denoising steps. In our setting with 100 steps for AnoF-Diff(Iterative) and 50 steps for Imdiffusion, AnoF-Diff (Parallel) achieves an $\sim$84.2× and 79.7×  acceleration compared to them.

We also demonstrate the online anomaly detection in the video. Due to the training data shift, instead of directly using the anomaly threshold calculated by the offline dataset, we collect a calibration dataset (three normal trajectories) from online and choose the maximum reconstruction loss as the calibrated abnormal threshold.


\section{Limitations}
\label{limitation}

Although this work demonstrates that our method is able to outperform baselines in anomaly detection, there are still limitations to our approach. First, the use of a fixed-length sliding window reduces the model’s ability to handle variable-length time series signals. The length of a sliding window is also an important hyperparameter that influences model performance.

Second, by sampling only local segments from the full trajectory, the sliding window approach inherently loses long-term dependency information, which can be critical for accurately modeling more complex time series behaviors. The correlation dependency of each window might also give important information for anomaly detection.

Finally, the data we used was gathered from task-specific controllers that were fairly simple. For a more complex control policy, we would expect a wider distribution of data and perhaps more subtle anomalies, and it is not yet clear if our method or any baseline will perform well under such conditions.

\section{CONCLUSIONS}

We present a method for multivariate time series anomaly detection designed for forceful tool use tasks.  Our model is based on a diffusion framework, with a proposed conditional masking mechanism applied specifically to the force-torque features, which can also generalize to other sensor features. We chose the four tool-use tasks to evaluate our method and compare it with state-of-the-art baselines. We also proposed parallel anomaly score evaluation, which can also be applied to other diffusion-based anomaly detection methods too.
Sec.\ref{limitation} discusses the limitations of our work. In future work, we would like to improve our model to address these limitations and explore integrating our model with a more sophisticated control strategy, such as anomaly recovery policies.

\addtolength{\textheight}{-12cm}   




\bibliographystyle{IEEEtran}
\bibliography{IEEEabrv,Bibliography}

\end{document}